\documentclass[final]{cvpr}

\usepackage{times}
\usepackage{epsfig}
\usepackage{graphicx}
\usepackage{amsmath}
\usepackage{amssymb}
\usepackage{bbm}

\usepackage[]{graphicx}
\usepackage{amsfonts, amssymb}
\usepackage{mdwmath}
\usepackage{mdwtab}
\usepackage[tight,footnotesize]{subfigure}
\usepackage{algorithm}	
\usepackage{algpseudocode}
\usepackage{balance}
\usepackage{comment}
\usepackage{gensymb}
\usepackage{multirow}
\usepackage{comment}
\usepackage[]{appendix}
\usepackage{mathrsfs}
\usepackage{footnote}
\makesavenoteenv{tabular}
\makesavenoteenv{table}

\DeclareMathOperator*{\argmax}{arg\,max}
\DeclareMathOperator*{\argmin}{arg\,min}

\usepackage[pagebackref=true,breaklinks=true,colorlinks,bookmarks=false]{hyperref}

\def\cvprPaperID{1822} 
\def\confYear{CVPR 2021}

\begin{document}

\title{DeepI2P: Image-to-Point Cloud Registration via Deep Classification}
\author{Jiaxin Li\\
Bytedance\\
\and
Gim Hee Lee\\
National University of Singapore\\
}
\maketitle

\begin{abstract}
This paper presents DeepI2P: a novel approach for cross-modality registration between an image and a point cloud. Given an image (e.g. from a rgb-camera) and a general point cloud (e.g. from a 3D Lidar scanner) captured at different locations in the same scene, our method estimates the relative rigid transformation between the coordinate frames of the camera and Lidar. Learning common feature descriptors to establish correspondences for the registration is inherently challenging due to the lack of appearance and geometric correlations across the two modalities. We circumvent the difficulty by converting the registration problem into a classification and inverse camera projection optimization problem. A classification neural network is designed to label whether the projection of each point in the point cloud is within or beyond the camera frustum. These labeled points are subsequently passed into a novel inverse camera projection solver to estimate the relative pose. Extensive experimental results on Oxford Robotcar and KITTI datasets demonstrate the feasibility of our approach. Our source code is available at \small \url{https://github.com/lijx10/DeepI2P}.

\end{abstract}

\section{Introduction}
Image-to-point cloud registration refers to the process of finding the rigid transformation, i.e., rotation and translation that aligns the projections of the 3D point cloud to the image. This process is equivalent to finding the pose, i.e., extrinsic parameters of the imaging device with respect to the reference frame of the 3D point cloud; and it has wide applications in many tasks in computer vision, robotics, augmented/virtual reality, etc. 

Although the direct and easy approach to solve the registration problem is to work with data from the same modality, i.e., image-to-image and point cloud-to-point cloud, several limitations exist in these same-modality registration approaches. 
For \textit{point cloud-to-point cloud} registration, it is impractical and costly to mount 
expensive and hard-to-maintain Lidars on large fleet of robots and mobile devices during operations. Furthermore, feature-based point cloud-to-point cloud registration \cite{deng2018ppf,zeng20173dmatch,li2019usip,yew20183dfeat} usually requires storage of $D$-dimensional features ($D\gg3$) in addition to the $(x,y,z)$ point coordinates, which increases the memory complexity. For \textit{image-to-image} registration, meticulous effort is required to perform SfM \cite{ullman1979interpretation,triggs1999bundle,fischler1981random} and store the image feature descriptors \cite{rublee2011orb,lowe1999object} corresponding to the reconstructed 3D points for feature matching. Additionally, image features are subjected to illumination conditions, seasonal changes, etc. Consequently, the image features stored in the map acquired in one season/time are hopeless for registration after a change in the season/time.

Cross-modality image-to-point cloud registration can be used to alleviate the aforementioned problems from the same modality registration methods. 
Specifically, a 3D point cloud-based map can be acquired once with Lidars, and then pose estimation can be deployed with images taken from cameras that are relatively low-maintenance and less costly on a large fleet of robots and mobile devices. Moreover, maps acquired directly with Lidars circumvents the hassle of SfM, and are largely invariant to seasonal/illumination changes. Despite the advantages of cross-modality image-to-point cloud registration, few research has been done due to its inherent difficulty. To the best of our knowledge, 2D3D-MatchNet \cite{feng20192d3d} is the only prior work on general image-to-point cloud registration. This work does cross-modal registration by learning to match image-based SIFT \cite{lowe1999object} to point cloud-based ISS \cite{zhong2009intrinsic} keypoints using deep metric-learning. However, the method suffers low inlier rate due to the drastic dissimilarity in the SIFT and ISS features across two modalities.
\begin{figure}[t] \centering 
{\includegraphics[width=0.23\textwidth]{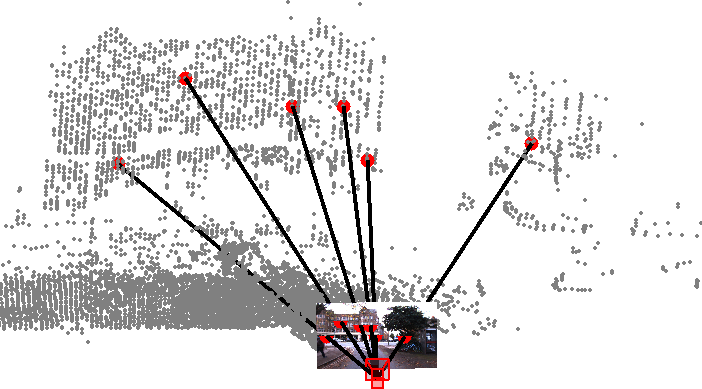}}
{\includegraphics[width=0.23\textwidth]{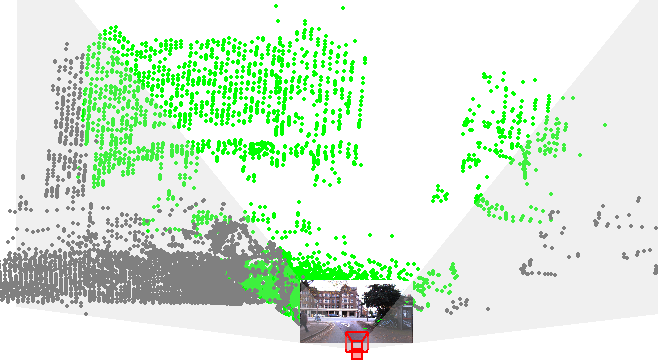}}
\vspace{-4pt}
\caption{Illustration of feature based registration on the left, e.g., 2D3D-MatchNet, and our feature-free DeepI2P on the right. Instead of detecting and matching features across modalities, we convert the registration problem into a classification problem.} \label{fig_3d_teaser}
\vspace{-8pt}
\end{figure}

In this paper, we propose the DeepI2P: a novel approach for cross-modal registration of an image and a point cloud \textit{without} explicit feature descriptors as illustrated in Fig.~\ref{fig_3d_teaser}. Our method requires lesser storage memory, i.e., $\mathcal{O}(3N)$ for the reference point cloud since we do not rely on feature descriptors to establish correspondences. Furthermore, the images captured by cameras can be directly utilized without SfM. We solve the cross-modal image-to-point cloud registration problem in two stages. In the first stage, we design a two-branch neural network that takes the image and point cloud as inputs, and outputs a label for every point that indicates whether the projection of this point is within or beyond the image frustum. 
The second stage is formulated as an unconstrained continuous optimization problem. The objective is to find the optimal camera pose, i.e., the rigid transformation with respect to the reference frame of the point cloud, such that 3D points labeled as within the camera frustum is correctly projected into the image. Standard solvers such as the Gauss-Newton algorithm can be used to solve our camera pose optimization problem. Extensive experimental results on the open-source Oxford Robotcar and KITTI datasets show the feasibility of our approach.   

The main contributions of this paper are listed as follow:
\begin{itemize}
    \item We circumvent the challenging need to learn cross-modal feature descriptor for registration by casting the problem into a two-stage classification and optimization framework. \vspace{-8pt}
    \item A two-branch neural network with attention modules to enhance cross-modality fusion is designed to learn labels of whether a 3D point is within or beyond the camera frustum. \vspace{-8pt}
    \item The inverse camera projection optimization is proposed to solve for the camera pose with the classification labels of the 3D points. \vspace{-8pt}
    \item Our method and the experimental results show a proof-of-concept that cross-modal registration can be achieved with deep classification. 
\end{itemize}

\section{Related Works}
\paragraph{Image-to-Image Registration.}
Images-to-image registrations \cite{shavit2019introduction, sattler2012improving} are done in the $\mathbb{P}^2$ space because of the lack of depth information. This is usually the first step to the computation of the projective transformation or SfM. Typical methods are usually based on feature matching. A set of features such as SIFT \cite{lowe1999object} or ORB \cite{rublee2011orb} are extracted from both source and target images. Correspondences are then established based on the extracted features, which can be used to solve for the rotation, translation using Bundle Adjustment \cite{triggs1999bundle,hartley2003multiple}, Perspective-n-Point solvers \cite{fischler1981random}, etc. Such techniques have been applied in modern SLAM systems \cite{engel2014lsd,mur2015orb,engel2017direct}. However, such methods are based on feature descriptors in the image modality to establish correspondences, and do not work for our general image-to-point cloud registration task.

\vspace{-3mm}
\paragraph{Point Cloud-to-Point Cloud Registration.}
The availability of 3D information enables direct registration between point clouds without establishing feature correspondences. Methods like ICP \cite{besl1992method,chen1992object}, NDT \cite{biber2003normal} work well with proper initial guess, and global optimization approaches such as Go-ICP \cite{yang2013go} work without initialization requirements. These methods are widely used in point cloud based SLAM algorithms like LOAM \cite{zhang2014loam}, Cartographer \cite{hess2016real}, etc. Recently data driven methods like DeepICP \cite{lu2019deepicp}, DeepClosestPoint \cite{wang2019deep}, RPM-Net \cite{yew2020-RPMNet}, etc, are also proposed. 
Although these approaches do not require feature correspondences, they still rely heavily on the geometrical details of the point structures in the same modality to work well. Consequently, these approaches cannot be applied to our task on cross-modal registration. 
Another group of common approaches is the two-step feature-based registration. Classical point cloud feature detectors \cite{tombari2013performance,zhong2009intrinsic,rusu20113d,dorai1997cosmos} and descriptors \cite{tombari2010unique,rusu2009fast} usually suffer from noise and clutter environments. Recently deep learning based feature detectors like USIP \cite{li2019usip}, 3DFeatNet \cite{yew20183dfeat}, and descriptors like 3DMatch \cite{zeng20173dmatch}, PPF-Net \cite{deng2018ppfnet}, PPF-FoldNet \cite{deng2018ppf}, PerfectMatch \cite{gojcic2019perfect}, have demonstrated improved performances in point cloud-based registration. Similar to image-to-image registration, these approaches require feature descriptors that are challenging to obtain in cross-modality registration.

\vspace{-3mm}
\paragraph{Image-to-Point Cloud Registration.}
To the best of our knowledge, 2D3D-MatchNet \cite{feng20192d3d} is the only prior work for general image-point cloud registration. It extracts images keypoints with SIFT \cite{lowe1999object}, and point cloud keypoints with ISS \cite{zhong2009intrinsic}. The image and point cloud patches around the keypoints are fed into each branch of a Siamese-like network and trained with triplet loss to extract cross-modal descriptors. At inference, it is a standard pipeline that consists of RANSAC-based descriptor matching and EPnP \cite{lepetit2009epnp} solver. Despite its greatly simplified experimental settings where the point clouds and images are captured at nearby timestamps with almost zero relative rotation, the low inlier rate of correspondences reveals the struggle for a deep network to learn common features across the drastically different modalities.
Another work \cite{yu2020monocular} establishes 2D-3D line correspondences between images and prior Lidar maps, but they requires accurate initialization, e.g., from a SLAM/Odometry system. In contrast, the general image-to-point cloud registration, including 2D3D-MatchNet \cite{feng20192d3d} and our DeepI2P do not rely on another accurate localization system. Some other works \cite{pham2020lcd,cattaneo2020global} focus on image-to-point cloud place recognition / retrieval without estimating the relative rotation and translation.


\section{Overview of DeepI2P}
We denote an image as $I \in \mathbb{R}^{3\times W \times H}$, where $W$ and $H$ are the image width and height, and a point cloud as $P=\{\mathbf{P}_1, \mathbf{P}_2, \cdots, \mathbf{P}_N \mid \mathbf{P}_n \in \mathbb{R}^3 \}$. The cross-modal image-to-point cloud registration problem is to solve for the rotation matrix $R \in \text{SO}(3)$ and translation vector $\mathbf{t} \in \mathbb{R}^{3}$ between the coordinate frames of the camera and point cloud. The problem is difficult because 
standard approaches such as ICP, PnP and Bundle Adjustment (BA) algorithms cannot be used due to the lack of point-to-pixel correspondences. 
Unlike the point cloud obtained from SfM, our point cloud is obtained from a point cloud scanner and does not contain any image-based feature descriptors. 
Establishing cross-modal point-to-pixel correspondence is non-trivial. This is because the points in the $\mathbb{R}^3$ space shares very little appearance and geometric correlations with the image in the $\mathbb{P}^2$ space. We circumvent the problem by designing our cross-modality image-to-point cloud registration approach to work without point-to-pixel correspondences. 

To this end, we propose a two-stage ``Frustum classification + Inverse camera projection'' pipeline. The first stage classifies each point in the point cloud into within or beyond the camera frustum. We call this the frustum classification, which is done easily by a deep network shown in Section~\ref{sec_classification}. In the second stage, we show that it is sufficient to solve the pose between camera and point cloud using only the frustum classification result. This is the inverse camera projection problem in Section~\ref{sec_inv_projection}. 
In our supplementary materials, we propose \textit{another} cross-modality registration method ``Grid classification + PnP'' as our baseline for experimental comparison. In the grid classification, the image is divided into a tessellation of smaller regular grids, and we predict the cell each 3D point projects into. The pose estimation problem can then be solved by applying RANSAC-based PnP to the grid classification output.

\section{Classification} \label{sec_classification}
The input to the network is a pair of image $I$ and point cloud $P$, and the output is a per-point classification for $P$. There are two classification branches: frustum and grid classification. The frustum classification assign a label to each point, $L^c=\{l^c_{1}, l^c_{2}, \cdots, l^c_{N}\}$, where $l^c_n \in \{0, 1\}$. $l^c_n=0$ if the point $\mathbf{P}_n$ is projected to outside the image $I$, and vice versa. Refer to the supplementary for the details of the grid classification branch used in our baseline.

\subsection{Our Network Design}
\begin{figure*}[t!] \centering
{\includegraphics[width=0.90\textwidth]{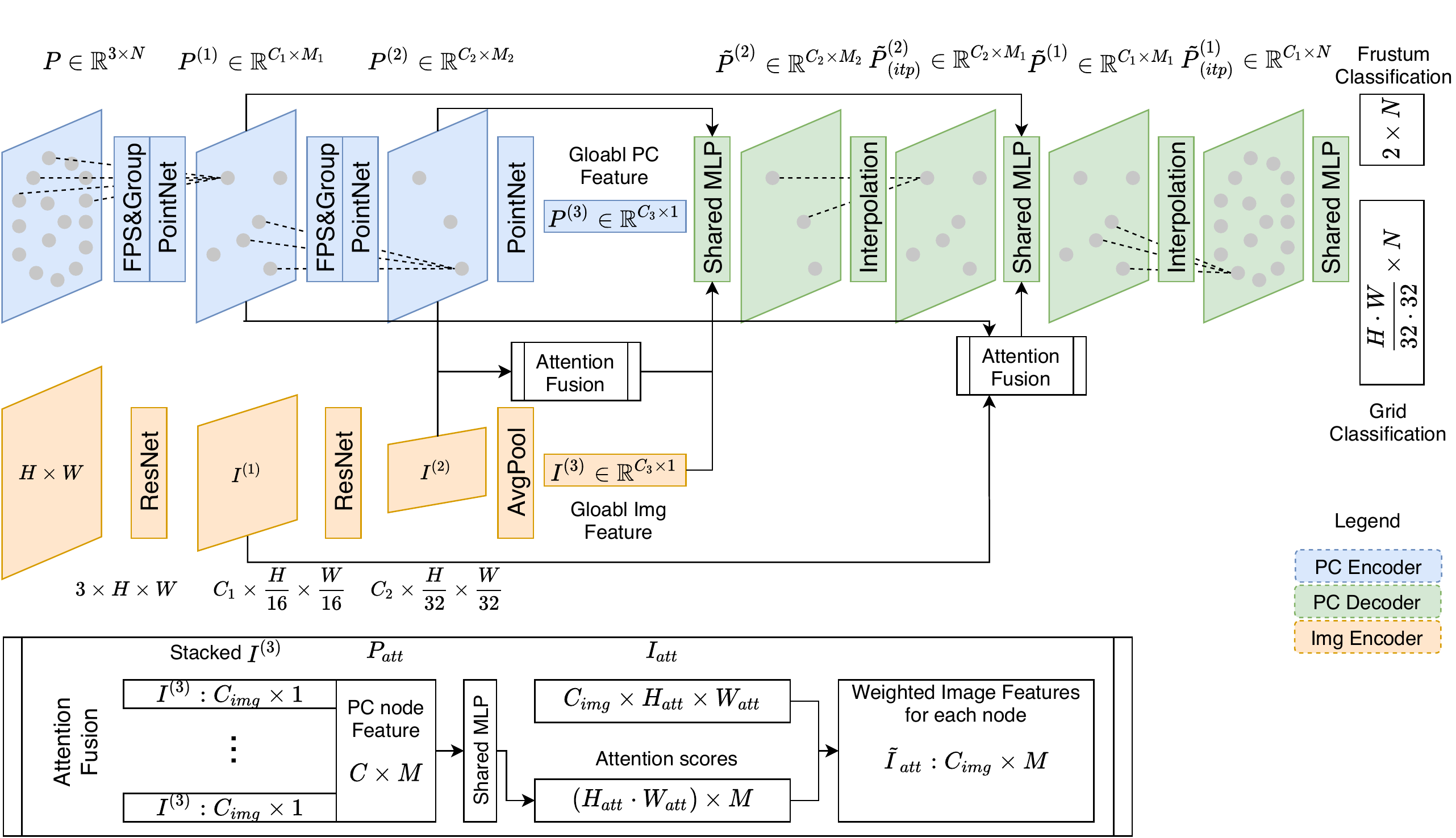}}
\vspace{-4pt}
\caption{Our network architecture for the classification problem.} \label{fig_network}
\vspace{-8pt}
\end{figure*}

As shown in Fig.~\ref{fig_network}, our per-point classification network consists of four parts: point cloud encoder, point cloud decoder, image encoder and image-point cloud attention fusion. The point cloud encoder/decoder follows the design of SO-Net \cite{li2018so} and PointNet++ \cite{qi2017pointnet++}, while the image encoder is a ResNet-34 \cite{he2016deep}. 
The classified points are then used in our inverse camera projection optimization in Section~\ref{sec_inv_projection} to solve for the unknown camera pose.

\vspace{-3mm}
\paragraph{Point Cloud Encoder.}
Given an input point cloud denoted as $P \in \mathbb{R}^{3\times N}$, a set of nodes $\mathfrak{P}^{(1)}\in \mathbb{R}^{3\times M_1}$ is sampled by Farthest Point Sampling (FPS). A point-to-node grouping \cite{li2018so} is performed to obtain $M_1$ clusters of points. Each cluster is processed by a PointNet \cite{qi2017pointnet} to get $M_1$ feature vectors of length $C_1$, respectively, i.e. $P^{(1)}\in\mathbb{R}^{C_1\times M_1}$. The point-to-node grouping is adaptive to the density of points. This is beneficial especially for point clouds from Lidar scans, where points are sparse at far range and dense at near range. The above sampling-grouping-PointNet operation is performed again to obtain another set of feature vectors $P^{(2)} \in \mathbb{R}^{C_2\times M_2}$. Finally, a PointNet is applied to obtain the global point cloud feature vector $P^{(3)}\in \mathbb{R}^{C_3\times 1}$.

\vspace{-3mm}
\paragraph{Image-Point Cloud Attention Fusion.}
The goal of the classification is to determine whether a point projects to the image plane (frustum classification) and which region it falls into (grid classification). Hence, it is intuitive that the classification requires fusion of information from both modalities. To this end, we design an Attention Fusion module to combine the image and point cloud information.
The input to the Attention Fusion module consists of three parts: a set of node features $P_{att}$ ($P^{(1)}$ or $P^{(2)}$), a set of image features $I_{att} \in \mathbb{R}^{C_{img}\times H_{att}\times W_{att}}$ ($I^{(1)}$ or $I^{(2)}$), and the global image feature vector $I^{(3)}$. As shown in Fig.~\ref{fig_network}, the image global feature is stacked and concatenated with the node features $P_{att}$, and fed into a shared MLP to get the attention score $S_{att} \in \mathbb{R}^{H_{att}W_{att}\times M}$. $S_{att}$ provides a weighting of the image features $I_{att}$ for $M$ nodes. The weighted image features are obtained by multiplying $I_{att}$ and $S_{att}$. The weighted image features can now be concatenated with the node features in the point cloud decoder.

\vspace{-3mm}
\paragraph{Point Cloud Decoder.}
The decoder takes the image and point cloud features as inputs, and outputs the per-point classification result. In general, it follows the interpolation idea of PointNet++ \cite{qi2017pointnet++}. At the beginning of the decoder, the global image feature $I^{(3)}$ and global point cloud feature $P^{(3)}$ are stacked $M_2$ times, so that they can be concatenated with the node features $P^{(2)}$ and the Attention Fusion output $\tilde{I}^{(2)}$. The concatenated $[I^{(3)}, \tilde{I}^{(2)}, P^{(3)}, P^{(2)}]$ is processed by a shared MLP to get $M_2$ feature vectors denoted as $\tilde{P}^{(2)} \in \mathbb{R}^{C_2\times M_2}$. We perform interpolation to get $\tilde{P}^{(2)}_{(itp)} \in \mathbb{R}^{C_2\times M_1}$, where the $M_2$ features are upsampled to $M_1 \geq M_2$ features. Note that $P^{(2)}$ and $\tilde{P}^{(2)}$ are associated with node coordinates $\mathfrak{P}^{(2)} \in \mathbb{R}^{3\times M_2}$. The interpolation is based on $k$-nearest neighbors between node coordinates $\mathfrak{P}^{(1)} \in \mathbb{R}^{3\times M_1}$, where $M_1 \geq M_2$. For each $C_2$ channel, the interpolation is denoted as:
\begin{equation} \label{eq_nn_interpolation}
    \tilde{P}^{(2)}_{(itp)_i} = \frac{\sum_{j=1}^k w_j \tilde{P}^{(2)}_j }{\sum_{j=1}^k w_j}, \text{\,\,where \,\,} w_j = \frac{1}{d(\mathfrak{P}^{(1)}_i, \mathfrak{P}^{(2)}_j)},
\end{equation}
and $\mathfrak{P}^{(2)}_j$ is one of the $k$-nearest neighbors of $\mathfrak{P}^{(1)}_i$ in $\mathfrak{P}^{(2)}$.
We get $\tilde{P}^{(2)}_{(itp)} \in \mathbb{R}^{C_2\times M_1}$
with the concatenate-sharedMLP-interpolation process. Similarly, we obtain $\tilde{P}^{(1)}_{(itp)} \in \mathbb{R}^{C_1\times N}$ after another round of operations. Lastly, we obtain the final output $(2+HW/(32\times32))\times N$, which can be reorganized into the frustum prediction scores $2\times N$ and grid prediction scores $(HW/(32\times32))\times N$. 

\subsection{Training Pipeline}
The generation of the frustum labels is simply a camera projection problem. During training, we are given the camera intrinsic matrix $K \in \mathbb{R}^{3\times 3}$ and the pose $G \in \text{SE}(4)$ between the camera and point cloud. The 3D transformation of a point $\mathbf{P}_i  \in \mathbb{R}^3$ from the point cloud coordinate frame to the camera coordinate frame is given by: 
\begin{equation} \label{equ_3d_point_transform}
\small
    \tilde{\mathbf{P}}'_i = [X'_i, Y'_i, Z'_i, 1]^\top = G\tilde{\mathbf{P}}_i = \begin{bmatrix} R & \mathbf{t} \\ 0 & 1\end{bmatrix} 
    \begin{bmatrix}\mathbf{P}_i \\ 1\end{bmatrix},
\end{equation}
and the point $\tilde{\mathbf{P}}'_i$ is projected into the image coordinate:
\begin{equation} \label{equ_cam_K_projection}
\small
    \tilde{\mathbf{p}}'_i = \begin{bmatrix}x'_i \\ y'_i \\ z'_i\end{bmatrix} = K\mathbf{P}'_i = \begin{bmatrix} f_x & 0 & c_x \\ 0 & f_y & c_y \\ 0 & 0 & 1 \end{bmatrix}\mathbf{P}'_i.
\end{equation}
Note that homogeneous coordinate is represented by a tilde symbol, e.g., $\tilde{\mathbf{P}}'_i$ is the homogeneous representation of $\mathbf{P}'_i$.
The inhomogeneous coordinate of the image point is:
\begin{equation} \label{equ_cam_projection_get_pixel}
    \mathbf{p}'_i = [p'_{x_i}, p'_{y_i}]^\top = [x'_i/z'_i, y'_i/z'_i]^\top.
\end{equation}

\vspace{-3mm}
\paragraph{Frustum Classification.}
For a given camera pose $G$, we define the function:
\begin{equation} \label{equ_camera_projection}
\small
\begin{split}
    &f(\mathbf{P}_i; G, K, H, W) = \\
    &\left\{ \begin{array}{ll}
                                    1 & : 0 \leq p'_{x_i} \leq W-1 \text{,\,} 0 \leq p'_{y_i} \leq H-1 \text{,\,} z'_i > 0 \\
                                    0 & : \text{otherwise}
                                  \end{array}, \right.
\end{split}
\end{equation}
which assigns a label of 1 to a point $\mathbf{P}_i$ that projects within the image, and 0 otherwise. 
Now the frustum classification labels are generated as $l^c_i = f(\mathbf{P}_i; G, K, H, W)$, where $G$ is known during training. 
In the Oxford Robotcar and KITTI datasets, we randomly select a pair of image and raw point cloud $(I, P_{raw})$, and compute the relative pose from the GPS/INS readings as the ground truth pose $G^p_c$.
We use $(I, P_{raw})$ with a relative distance within a specified interval in our training data.
However, we observe that the rotations in $G^p_c$ are close to zero from the two datasets since the cars used to collect the data are mostly undergoing pure translations. To avoid overfitting to such scenario, we apply
randomly generated rotations $G_r$ onto the raw point cloud to get the final point cloud $P=G_rP_{raw}$ in the training data. Furthermore, the ground truth pose is now given by $G = G^p_c G_r^{-1}$. 
Note that random translation can also be included in $G_r$, but it does not have any effect on the training since the network is translational equivariant.

\vspace{-3mm}
\paragraph{Training Procedure.}
The frustum classification training procedure is summarized as:
\begin{enumerate}
    \item Select a pair of image and point cloud $(I, P_{raw})$ with relative pose $G^p_c$.
    \vspace{-8pt}
    \item Generate 3D random transformation $G_r$, and apply it to get $P=G_rP_{raw}$ and $G=G_c^pG_r^{-1}$.
    \vspace{-8pt}
    \item Get the ground truth per-point frustum labels $l^c_i$ according to Eq.~\ref{equ_camera_projection}.
    \vspace{-8pt}
    \item Feed $({I, P})$ into the network illustrated in Fig.~\ref{fig_network}.
    \vspace{-8pt}
    \item Frustum prediction $\hat{L}^c=\{\hat{l}^c_1, \cdots, \hat{l}^c_n\}, \hat{l}^c_i \in \{0, 1\}$.
    \vspace{-8pt}
    \item Apply cross entropy loss for the classification tasks to train the network.
\end{enumerate}
\section{Pose Optimization} \label{sec_optimization}
We now formulate an optimization method to get the pose of the camera in the point cloud reference frame with the frustum classification results.
Note that we do not use deep learning in this step since the physics and geometry of the camera projection model is already well-established.

Formally, the pose optimization problem is to solve for $\hat{G}$, given the point cloud $P$, frustum predictions $\hat{L}^c=\{\hat{l}^c_1, \cdots, \hat{l}^c_N\}, \hat{l}^c_i \in \{0, 1\}$, and camera intrinsic matrix $K$. In this section, we describe our inverse camera projection solver to solve for $\hat{G}$.


\subsection{Inverse Camera Projection} \label{sec_inv_projection}
The frustum classification of a point, i.e., $L^c$ given $G$ defined in Eq.~\ref{equ_camera_projection} is based on the forward projection of a camera. 
The inverse camera projection problem is the other way around, i.e, determine the optimal pose $\hat{G}$ that satisfies a given $\hat{L}^c$. It can be written more formally as:
\begin{equation} \label{equ_inverse_cam_proj_original}
\small
    \hat{G} = \argmax_{G\in\text{SE}(3)} \sum_{i=1}^{N} \big(f(\mathbf{P}_i; G, K, H, W) - 0.5\big) \big(\hat{l}^c_i - 0.5\big).
\end{equation}
Intuitively, we seek to find the optimal pose $\hat{G}$ such that all 3D points with label $\hat{l}^c_i=1$ from the network are projected into the image, and vice versa. However, a naive search of the optimal pose in the SE$(3)$ space is intractable. To mitigate this problem, we relax the cost as a function of the distance from the projection of a point to the image boundary, i.e., a $H\times W$ rectangle. 

\vspace{-3mm} \paragraph{Frustum Prediction Equals to 1.}
Let us consider a point $\mathbf{P}_i$ with the prediction $\hat{l}^c_i=1$. We define cost function:
\begin{equation} \label{equ_cost_x_in_image}
    g(p'_{x_i}; W) = \text{max}(-p'_{x_i}, 0) + \text{max}(p'_{x_i} - W, 0)
\end{equation}
that penalizes a pose $G$ which causes $p'_{x_i}$ of the projected point $\mathbf{p}'_i=[p'_{x_i}, p'_{y_i}]^\top$ (c.f. Eq.~\ref{equ_cam_projection_get_pixel}) to fall outside the borders of the image width.
Specifically, the cost is zero when $p'_{x_i}$ is within the image width, and negatively proportional to the distance to the closest border along the image x-axis otherwise. A cost $g(p'_{y_i}; H)$ can be analogously defined along image y-axis.
In addition, cost function $h(\cdot)$ is defined to avoid the ambiguity of $\mathbf{P}'_i$ falling behind the camera:
\begin{equation}
    h(z'_i) = \alpha \cdot \text{max}(-z', 0),
\end{equation}
where $\alpha$ is a hyper-parameter that balances the weighting between $g(\cdot)$ and $h(\cdot)$.

\vspace{-3mm} \paragraph{Frustum Prediction Equals to 0.}
We now consider a point $\mathbf{P}_i$ with prediction $\hat{l}^c_i=0$. The cost defined along the image x-axis is given by:
\begin{equation} \label{equ_cost_x_out_image}
    u(p'_{x_i}; W) = \frac{W}{2} - \bigg|p'_{x_i} - \frac{W}{2}\bigg|. 
\end{equation}
It is negative when $p'_{x_i}$ falls outside the borders along the image width, and positively proportional to the distance to the closest border along the image x-axis otherwise. Similarly, an analogous cost $u(p'_{y_i}; H)$ along the y-axis can be defined. 
Furthermore, an indicator function:  
\begin{equation}
\begin{split}
    &\mathbbm{1}(p'_{x_i}, p'_{y_i}, z'_i; H, W) = \frac{\text{max}(u(p'_{x_i}; W), 0)}{u(p'_{x_i}; W)} \\
    &\quad \quad \quad \quad \quad \quad \cdot \frac{\text{max}(u(p'_{y_i}; H), 0)}{u(p'_{y_i}; H)} \cdot \frac{\text{max}(z'_i, 0)}{z'_i}
\end{split}
\end{equation}
is required to achieve the target of zero cost when $\mathbf{p}'_i$ is outside the $H\times W$ image or $\mathbf{P}'_i$ is behind the camera (i.e. $z'_i<0$).

\vspace{-3mm} \paragraph{Cost Function.}
Finally, the cost function for a single point $\mathbf{P}_i$ is given by:
\begin{equation} \label{equ_element_cost_function_r}
    r_i(G ; \hat{l}^c_i) = 
    \left\{ \begin{array}{ll}
        r_i^0 & : \hat{l}^c_i = 0 \\
        r_i^1 & : \hat{l}^c_i = 1
    \end{array}\right., \quad \text{where}\\
\end{equation}    
\begin{equation*}
\begin{split}
    & r_i^0 = (u(p'_{x_i}; W) + u(p'_{y_i}; H))\cdot \mathbbm{1}(p'_{x_i}, p'_{y_i}, z'_i; H, W), \\
    & r_i^1 = g(p'_{x_i}; W) + g(p'_{y_i}; H) + h(z'_i).
\end{split}
\end{equation*}
$p'_{x_i}, p'_{y_i}, z_i$ are functions of $G$ according to Eq.~\ref{equ_3d_point_transform}, \ref{equ_cam_K_projection} and \ref{equ_cam_projection_get_pixel}. Image height $H$, width $W$ and camera intrinsics $K$ are known. Now the optimization problem in Eq.~\ref{equ_inverse_cam_proj_original} becomes:
\begin{equation} \label{equ_relaxed_cost_function}
    \hat{G} = \argmin_{G\in \text{SE}(3)} \sum_{i=1}^{n} r_i(G; \hat{l}^c_i)^2.
\end{equation}
This is a typical unconstrained least squares optimization problem. We need proper parameterization of the unknown transformation matrix,
\begin{equation}
\small
    G = \begin{bmatrix} R & \mathbf{t} \\ \mathbf{0} & 1 \end{bmatrix}, \text{\,\, with \,\,} R \in SO(3), \mathbf{t} \in \mathbb{R}^3,
\end{equation}
where $G \in SE(3)$ is an over-parameterization that can cause problems in the unconstrained continuous optimization. To this end, we use the Lie-algebra representation $\xi \in \mathfrak{se}(3)$ for the minimal parameterization of $G\in \text{SE}(3)$. The exponential map $G = \text{exp}_{\mathfrak{se}(3)}(\xi)$ converts $\mathfrak{se}(3) \mapsto \text{SE}(3)$, while the log map
$\xi = \text{log}_{SE(3)}(G)$ converts $\text{SE}(3) \mapsto \mathfrak{se}(3)$. Similar to \cite{engel2014lsd}, we define the $\mathfrak{se}(3)$ concatenation operator $\circ: \mathfrak{se}(3) \times \mathfrak{se}(3) \mapsto \mathfrak{se}(3)$ as:
\begin{equation}
\small
    \xi_{ki} = \xi_{kj} \circ \xi_{ji} = \text{log}_{\text{SE}(3)}\big( \text{exp}_{\mathfrak{se}(3)}(\xi_{kj}) \cdot \text{exp}_{\mathfrak{se}(3)}(\xi_{ji}) \big),
\end{equation}
and the cost function in Eq.~\ref{equ_relaxed_cost_function} can be re-written
with the proper exponential or log map modifications into:
\begin{equation} \label{equ_relaxed_cost_function_lie_algebra}
\begin{split}
    & \hat{G} = \text{exp}_{\mathfrak{se}(3)}(\hat{\xi}), \quad \text{where}\\ 
    & \hat{\xi} = \argmin_{\xi} \|\mathbf{r}\|^2 = \argmin_{\xi} \sum_{i=1}^{n} r_i(\xi; \hat{l}^c_i)^2.
\end{split}
\end{equation}

\vspace{-6mm} \paragraph{Gauss-Newton Optimization.} 
\begin{figure*}[t] \centering 
{\includegraphics[width=0.3\textwidth]{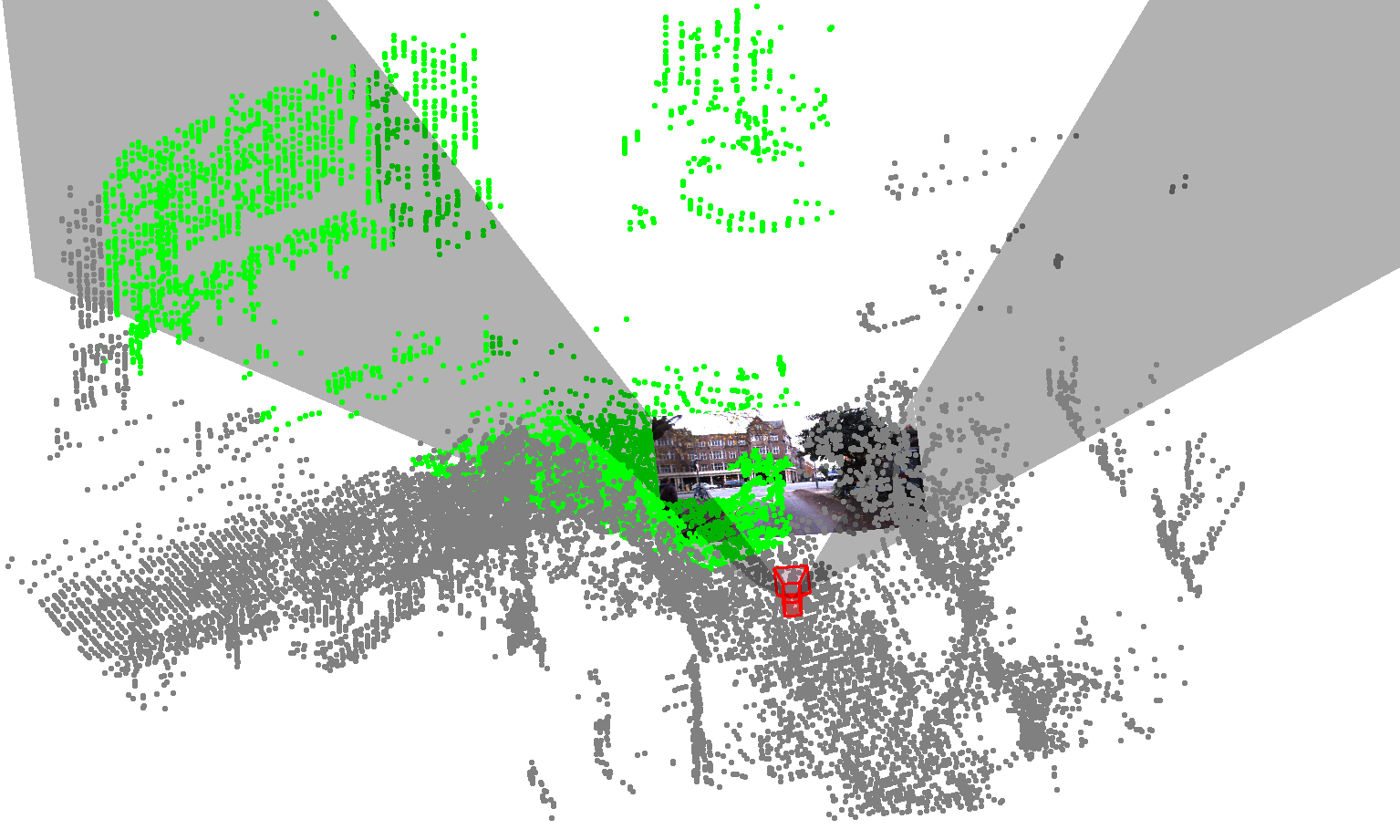}}
{\includegraphics[width=0.3\textwidth]{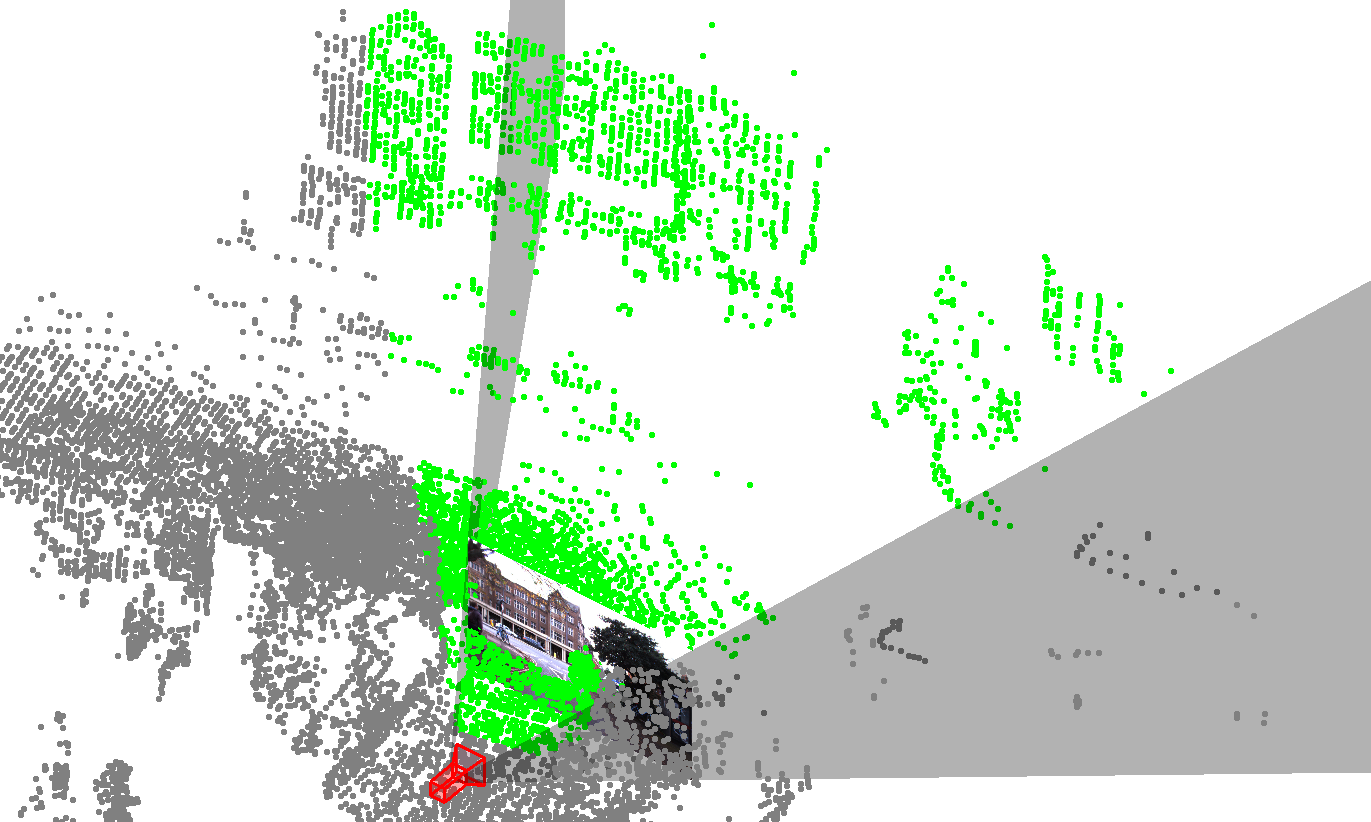}}
{\includegraphics[width=0.3\textwidth]{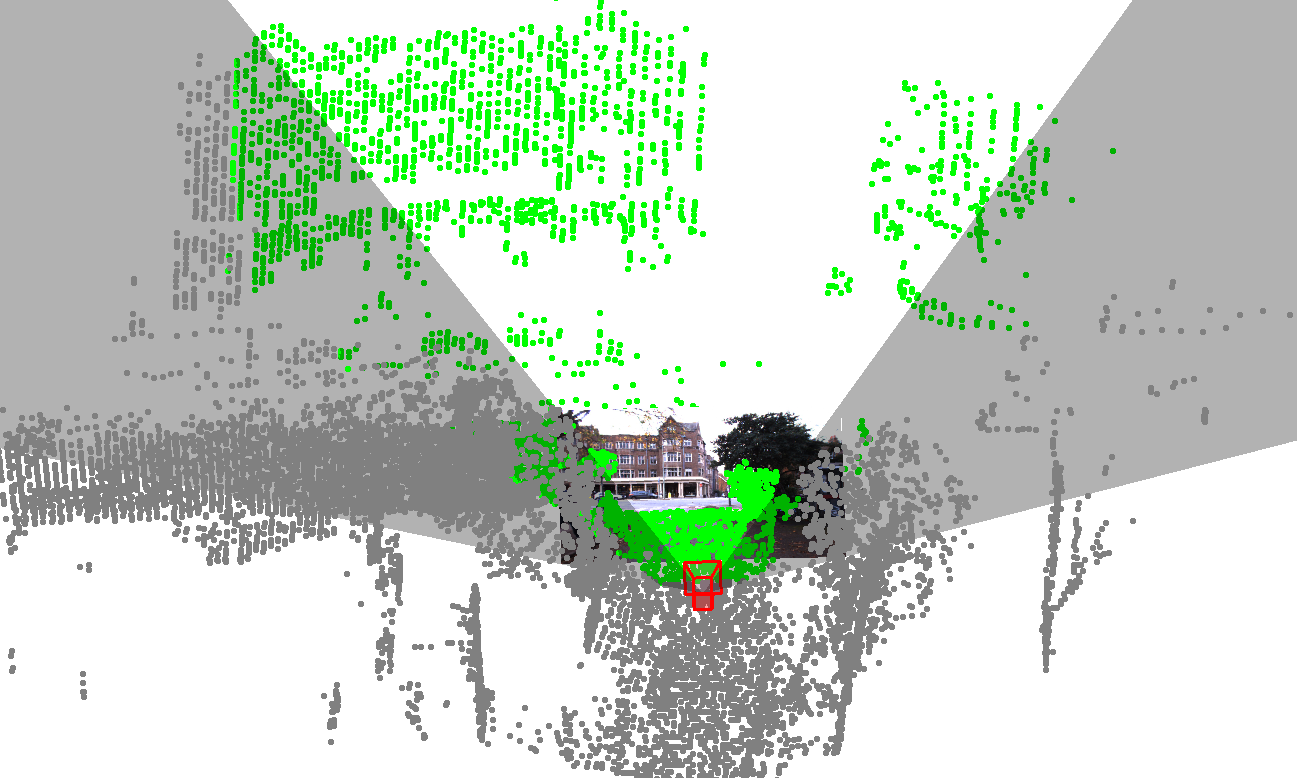}}
\vspace{-2pt}
\caption{Visualizations of the Gauss-Newton at iteration 0 / 40 / 80 from left to right. Green points are classified as inside image FoV. } \label{fig_gauss_newton_optimization}
\vspace{-4pt}
\end{figure*}

Eq.~\ref{equ_relaxed_cost_function_lie_algebra} is a typical least squares optimization problem that can be solved by the Gauss-Newton method. During iteration $i$ with the current solution $\xi^{(i)}$, the increment $\delta \xi ^{(i)}$ is estimated by a Gauss-Newton second-order approximation:
\begin{equation}
\small
    \delta \xi^{(i)} = -(J^\top J)^{-1}J^\top\mathbf{r}(\xi^{(i)}), \text{\,\,where\,\,} J=\frac{\partial \mathbf{r}(\epsilon\circ \xi^{(i)})}{\partial \epsilon}\bigg|_{\epsilon=0}, 
\end{equation}
and the update is given by $\xi^{(i+1)} = \delta \xi^{(i)} \circ \xi^{(i)}$. Finally the inverse camera projection problem is solved by performing the exponential map $\hat{G} = \text{exp}_{\mathfrak{se}(3)}(\hat{\xi})$. A visualization of our iterative optimization is presented in Fig.~\ref{fig_gauss_newton_optimization}.

\section{Experiments}
Our image-to-point cloud registration approach is evaluated with Oxford Robotcar \cite{maddern20171} and KITTI \cite{geiger2013vision} dataset. 

\vspace{-3mm} 
\paragraph{Oxford dataset.} The point clouds are built from the accumulation of the 2D scans from a 2D Lidar. Each point cloud is set at the size of radius $50$m, i.e. diameter $100$m. The images are captured by the center camera of a Bumblebee tri-camera rig. Similar to 3DFeat-Net \cite{yew20183dfeat}, 35 traversals are used for training, while 5 traversals are for testing. 
In training and inference, the image-point cloud pair is selected using the following steps: a) Choose a Lidar point cloud from one of the traversals. b) Randomly select an image from the same traversal and captured within $\pm 10$m from the coordinate origin of the point cloud. The relative pose of the camera to the point cloud coordinate frame is $G^p_c$. c) Apply a random 2D rotation (around the up-axis) and translation (along the x-y plane) $G_r$ to the point cloud. d) The objective is to recover the ground truth transformation $G_{gt}=G^p_cG_r^{-1}$. There are 130,078 point clouds for training and 19,156 for testing. 

\vspace{-3mm} 
\paragraph{KITTI Odometry dataset.} Point clouds are directly acquired from a 3D Lidar. The image-point cloud pairs are selected in a similar way to that in Oxford dataset, i.e., a pair of image and point cloud is captured within $\pm10$m. The images are from both the left and right cameras that are facing the front. 
We follow the common practice of utilizing the 0-8 sequences for training, and 9-10 for testing. In total there are 20,409 point clouds for training, and 2,792 for testing. 

\vspace{1mm}
\noindent \textbf{Remarks:} Note that a KITTI point cloud is from a single frame 3D Lidar scan, while an Oxford point cloud is an accumulation of 2D Lidar scans over $100$m. As a result, point clouds in KITTI suffers from severe occlusion, sparse measurement at far range, etc. More details of the two dataset configurations are in the supplementary materials.

\subsection{Implementation Details}
\paragraph{Classification Network.}
Refer to our supplementary materials for the network implementation details. These includes
parameters of the PointNet and SharedMLP modules, number of nodes in $\mathfrak{P}^{(1)}, \mathfrak{P}^{(2)}$, number of nearest neighbor $k$ for point cloud interpolation in the decoder, etc.

\vspace{-3mm} \paragraph{Inverse Camera Projection.} The initial guess $G^{(0)}$ in our proposed inverse camera projection (c.f. Section~\ref{sec_inv_projection}) is critical since the solver for   Eq.~\ref{equ_relaxed_cost_function_lie_algebra} is an iterative approach. To alleviate the initialization problem, we perform the optimization $60$ times with randomly generated initialization $G^{(0)}$, and select the solution with the lowest cost. 
In addition, the 6DoF search space is too large for random initialization. We mitigate this problem by leveraging on the fact that our datasets are from ground vehicles to perform random initialization in 2D instead. Specifically, $R^{(0)}$ is initialized as a random rotation around the up-axis, and $\mathbf{t}^{(0)}$ as a random translation in the x-y horizontal plane. Our algorithm is implemented with Ceres \cite{ceres-solver}.

\subsection{Registration Accuracy}
\begin{table*}[t!]
\centering
\caption{Registration accuracy on the Oxford and KITTI datasets.}
\label{tbl_reg_accuracy}
{%
\begin{tabular}{l|c|c|c|c}
\hline
                    & \multicolumn{2}{c|}{Oxford} & \multicolumn{2}{c}{KITTI} \\ \cline{2-5}
                    & RTE (m)     & RRE (\degree)      & RTE (m)    & RRE (\degree)     \\ \hline
Direct Regression   & $5.02\pm2.89$  & $10.45\pm16.03$  & $4.94\pm2.87$ & $21.98\pm31.97$ \\ 
MonoDepth2 \cite{godard2019digging} + USIP \cite{li2019usip}  & $33.2\pm46.1$ & $142.5\pm139.5$ & $30.4\pm42.9$ & $140.6\pm157.8$ \\
MonoDepth2 \cite{godard2019digging} + GT-ICP & $\bf{1.3\pm1.5}$  & $6.4\pm7.2$ & $\bf{2.9\pm2.5}$ & $12.4\pm10.3$ \\
2D3D-MatchNet \cite{feng20192d3d} (No Rot$^\S$)     & $1.41$  & $6.40$ & NA & NA \\ \hline
Grid Cls. + PnP          & $1.91\pm1.56$  & $5.94\pm10.72$   & $3.22\pm3.58$ & $10.15\pm13.74$ \\
Frus. Cls. + Inv.Proj. 3D & $2.27\pm2.19$  & $15.00\pm13.64$  & $3.17\pm3.22$ & $15.52\pm12.73$ \\
Frus. Cls. + Inv.Proj. 2D & $1.65\pm1.36$  & $\bf{4.14\pm4.90}$    & $3.28\pm3.09$ & $\bf{7.56\pm7.63}$   \\ \hline
\end{tabular}%
}
\newline
{$^\S$\small{Point clouds are not randomly rotated in the experiment setting of 2D3D-MatchNet \cite{feng20192d3d}.} \par}
\vspace{-8pt}
\end{table*}
%
%

\begin{figure}[t!] \centering 
{\includegraphics[width=0.115\textwidth]{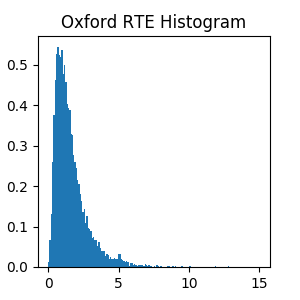}}
{\includegraphics[width=0.115\textwidth]{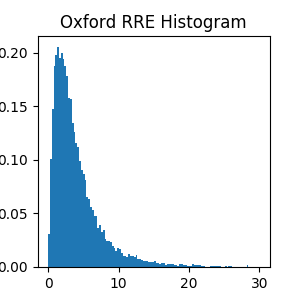}}
{\includegraphics[width=0.115\textwidth]{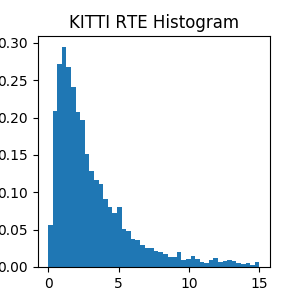}}
{\includegraphics[width=0.115\textwidth]{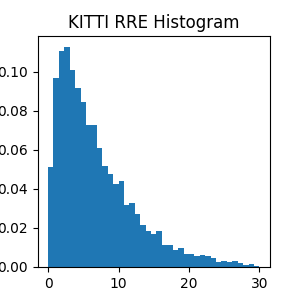}}
\vspace{-8pt}
\caption{Histograms of image-point cloud registration RTE and RRE on the Oxford and KITTI datasets. x-axis is RTE (m) and RRE (\degree), and y-axis is the percentage.} \label{fig_histogram}
\vspace{-12pt}
\end{figure}

The frustum classification accuracy is 98\% and 94\% on the Oxford and KITTI dataset, respectively. However, these numbers does not translate directly to the registration accuracy. Following the practice of \cite{li2019usip,yew20183dfeat}, the registration is evaluated with two criteria: average Relative Translational Error (RTE) and average Relative Rotation Error (RRE). The results are shown in Table~\ref{tbl_reg_accuracy} and Fig.~\ref{fig_histogram}. 

\textit{Grid Cls. + PnP } is the result of our ``Grid classification + PnP'' baseline method (see Supplementary materials for details). The RANSAC PnP algorithm optimizes the full 6-DoF $\hat{G}$ without any constraints. \textit{Frus. Cls. + Inv.Proj.} represents the result of our ``Frustum classification + Inverse camera projection'' method. The difference between \textit{Frus. Cls. + Inv.Proj. 3D} and \textit{Frus. Cls. + Inv.Proj. 2D} is that the former is optimizing the full 6-DoF $\hat{G}$, while the latter constrains $\hat{G}$ to be 3-DOF, i.e., translation on x-y horizontal plane and rotation around the up-axis.

Due to the lack of existing approaches in solving the image-to-point cloud registration problem under the same setting, we further compare our DeepI2P with 4 other approaches that may contain unfair advantages over ours in their input data modality or configurations. 

\vspace{2mm}
\noindent \textbf{1) Direct Regression} uses a deep network to directly regress the relative poses. It consists of the Point Cloud Encoder and Image Encoder in Section~\ref{sec_classification}. The global point cloud feature and global image feature are concatenated into a single vector and processed by a MLP that directly regresses $\hat{G}$. See the supplementary materials for more details of this method. Table \ref{tbl_reg_accuracy} shows that our DeepI2P significantly outperforms the simple regression method.

\vspace{2mm}
\noindent \textbf{2) Monodepth2+USIP} converts the cross-modality registration problem into point cloud-based registration by using Monodepth2 \cite{godard2019digging} to estimate a depth map from a single image. The Lidar point cloud is used to calibrate the scale of depth map from MonoDepth2, i.e. the scale of the depth map is perfect. Subsequently, the poses between the depth map and point cloud are estimated with USIP \cite{li2019usip}. This is akin to same modality point cloud-to-point cloud registration. Nonetheless, Table \ref{tbl_reg_accuracy} shows that this approach under-performs. 
This is probably because the depth map is inaccurate and USIP does not generalize well on depth maps.

\vspace{2mm}
\noindent \textbf{3) Monodepth2+GT-ICP} acquires a depth map with absolute scale in the same way as Monodepth2+USIP. However, it uses Iterative Closest Point (ICP) \cite{besl1992method,chen1992object} to estimate the pose between the depth map and point cloud. Note that ICP fails without proper initialization, and thus we use the ground truth (GT) relative pose for initialization. Table \ref{tbl_reg_accuracy} shows that our DeepI2P achieves similar RTE and better RRE compared to Monodepth2+GT-ICP, despite the latter has the unfair advantages of ground truth initialization and the depth map is perfectly calibrated.

\vspace{2mm}
\noindent \textbf{4)2D3D-MatchNet \cite{feng20192d3d}} is the only prior work for cross-modal image-to-point cloud registration to our best knowledge. However, the rotation between camera and Lidar is almost zero in their experiment setting. This is because the images and point clouds are taken from temporally consecutive timestamps without additional augmentation. In contrast, the point clouds in our experiments are always randomly rotated. This means 2D3D-MatchNet is solving a much easier problem, but their results are worse than ours.

\vspace{-3mm}
\paragraph{Distribution of Errors.}
The distribution of the registration RTE (m) and RRE (\degree) on the Oxford and KITTI dataset are shown in Fig.~\ref{fig_histogram}. It can be seen that our performance is better on Oxford than KITTI.
Specifically, the mode of the translational/rotational errors are  $\sim 1.5\text{m} / 3\degree$ on Oxford and $\sim 2\text{m} / 5\degree$ on KITTI. The translational/rotational error variances are also smaller on Oxford. 


\vspace{-3mm}
\paragraph{Acceptance of RTE/RRE.}
There are other same-modality methods that solves registration on Oxford and KITTI dataset, e.g. USIP \cite{li2019usip} and 3DFeatNet \cite{yew20183dfeat} that gives much better RTE and RRE. These methods work only on point cloud-to-point cloud instead of image-to-point cloud data, and thus are not directly comparable to our method. Furthermore, we note that the reported accuracy of our DeepI2P in Table \ref{tbl_reg_accuracy} is sufficient for non life-critical applications such as frustum localization of mobile devices in both indoor and outdoor environments. 

\vspace{-3mm}
\paragraph{Oxford vs KITTI.} 
Our performance on Oxford is better than on KITTI for several reasons: 1) The point clouds in Oxford are built from 2D scans accumulated over $100$m from a Lidar scanner, while KITTI point cloud is a single scan from a 3D Lidar as shown in Fig.~\ref{fig_3d_vis}. Consequently, the occlusion effect in KITTI is severe. For example, the image captured in timestamp $t_j$ is seeing things that are mostly occluded/unobserved from the point cloud of timestamp $t_i$. Given the limited Field-of-View (FoV) of the camera, the cross-modality registration becomes extremely challenging since two modalities are observing different scene contents. 2) The vertical field-of-view of the point clouds in KITTI is very limited, which leads to lack of distinctive vertical structures for cross-modality matching. Most of the points on the ground are structureless and thus not useful for matching across modalities. 3) ``KITTI Odometry" is a small dataset that contains only 20,409 point clouds for training, while Oxford is $\sim 6.4\times$ larger with 130,078 point clouds. As a result, we observe severe network overfitting in the KITTI dataset but not in the Oxford dataset.

\subsection{Ablation Study}
\paragraph{Initialization of our Gaussian-Newton and Monodepth2+ICP.} 
In the 2D registration setting, there are 3 unknown parameters - rotation $\theta$, translation $t_x, t_y$. For our method, the initial $\theta$ is easily obtained by aligning the average yaw-angles of the predicted in-frustum points with the camera principle axis. Therefore, our 60-fold initialization is only for two-dimensional search for $t_x, t_y$. In contrast, Monodepth2+ICP requires three-dimensional search for $\theta, t_x, t_y$. As shown in Tab.~\ref{tab_ablation}, our DeepI2P is robust to initialization, while Monodepth2+ICP performs a lot worse with 60-fold random initialization. 

\vspace{-8pt}
\paragraph{In-image occlusion.} In Oxford/Kitti, occlusion effect can be significant when the translation is large, e.g. $>5$m. Tab.~\ref{tab_ablation} shows that the registration improves when the maximum translation limit decrease ($15$m $\to 10$m $\to 5$m).

\vspace{-8pt}
\paragraph{Cross-attention module.} Tab.~\ref{tab_ablation} shows the significant drop in registration accuracy w/o cross-attention module. Additionally, we also see the coarse classification accuracy drops significantly from 98\% to 80\%.

\vspace{-8pt}
\paragraph{3D point density.} As shown in Tab.~\ref{tab_ablation}, our registration accuracy decreases with reducing point density. Nonetheless, the performance drop is reasonable even when the point density drops to 1/4 ($20480 \to 5120$).

\begin{table}[h]
\centering
\footnotesize
\caption{Registration accuracy on Oxford}
\label{tab_ablation}
\resizebox{0.48\textwidth}{!}{%
\setlength\tabcolsep{3.0pt} 
\begin{tabular}{l|ccc|cc}
\hline
                 & \# points & \# init. & $t$ limit & RTE  & RRE  \\ \hline
DeepI2P          & 20480     & 60       & 10      & 1.65 & 4.14 \\ \hline
DeepI2P          & 20480     & 30       & 10      & 1.81 & 4.37     \\
DeepI2P          & 20480     & 10       & 10      & 2.00 & 4.56      \\
DeepI2P          & 20480     & 1        & 10      & 3.52 & 5.34      \\
DeepI2P          & 20480     & 60       & 5       & 1.52 & 3.30     \\
DeepI2P          & 20480     & 60       & 15      & 1.96 & 4.74     \\
DeepI2P          & 10240     & 60       & 10      & 1.80 & 4.35     \\
DeepI2P          & 5120      & 60       & 10      & 1.94 & 4.63     \\
DeepI2P noAtten. & 20480     & 60       & 10      & 6.88 & 20.93      \\
MonoDepth2+ICP   & 20480     & 60       & 10      & 8.45 & 75.54 \\ \hline
\end{tabular}%
}
\vspace{-2mm}
\end{table}

\subsection{Visualizations}
\begin{figure}[t!] \centering 
{\includegraphics[width=0.49\textwidth]{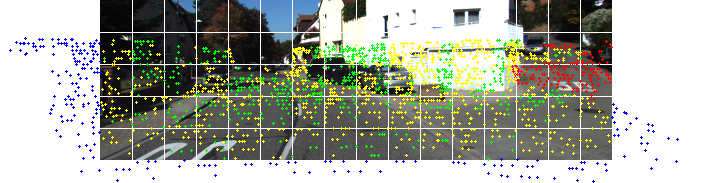}}
\caption{Visualization of the frustum and grid classification results projected onto the images. Green - both frustum and grid predictions are correct. Yellow - frustum prediction is correct but grid prediction is wrong. Red - frustum prediction is outside image FoV, but ground truth label is inside FoV. Blue - frustum prediction is inside image FoV, but ground truth label is outside FoV. Best view in color and zoom-in.} \label{fig_cls_vis_in_image}
\vspace{-4pt}
\end{figure}

\begin{figure}[t] \centering 
{\includegraphics[width=0.23\textwidth]{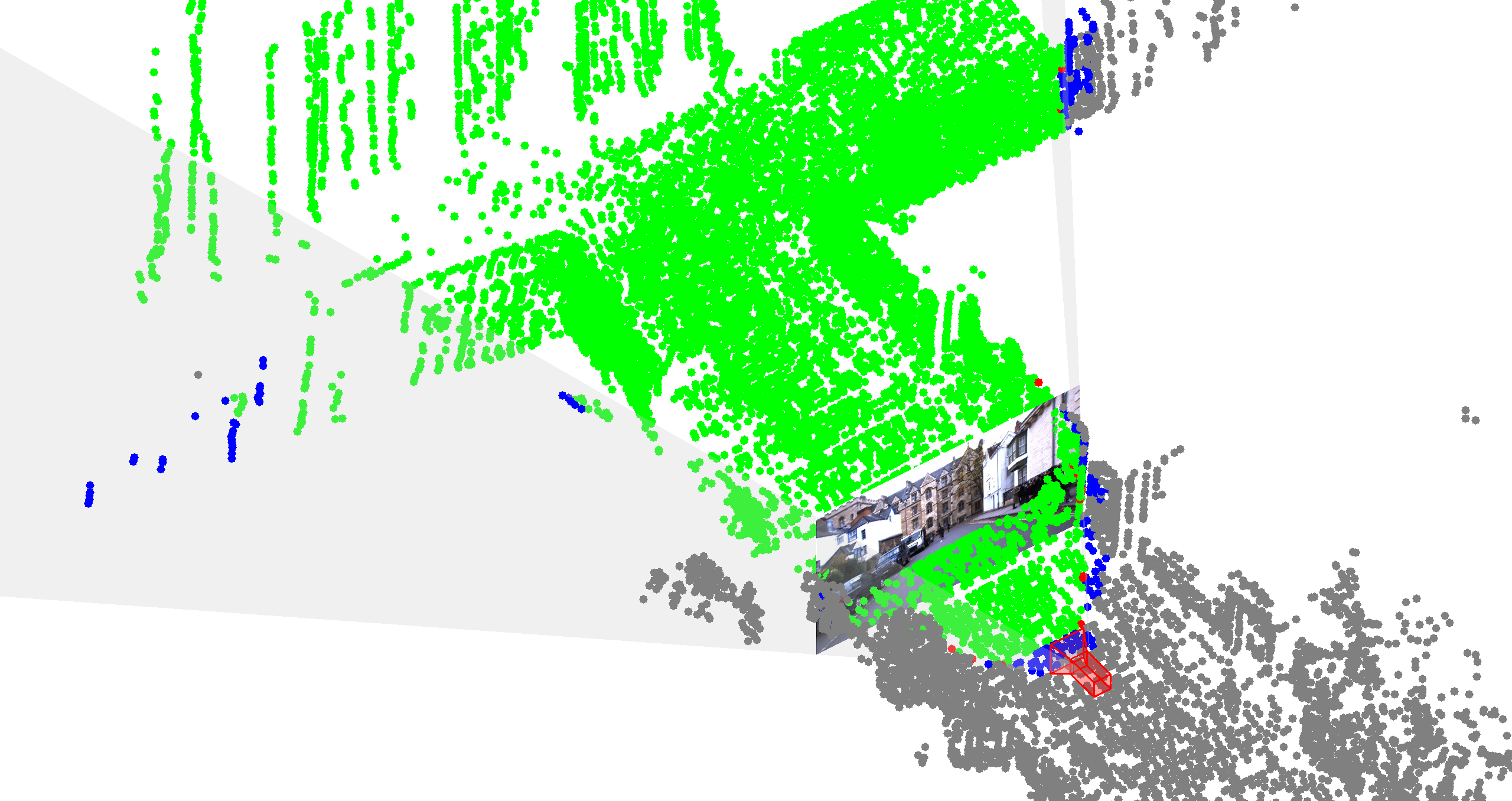}}
{\includegraphics[width=0.23\textwidth]{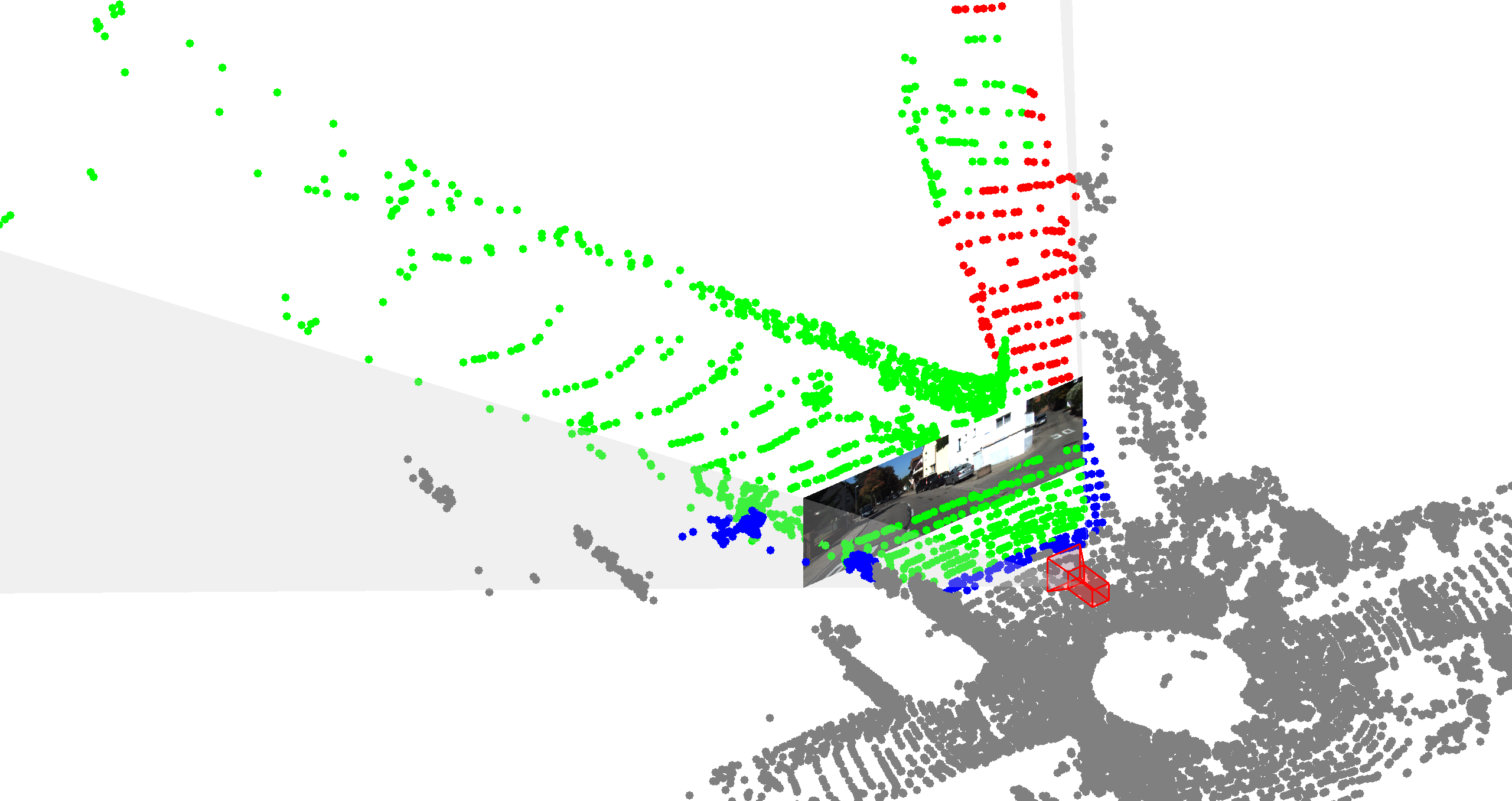}}
\caption{3D Visualization of the frustum classification and inverse camera projection on the Oxford (Left) and KITTI (Right).} \label{fig_3d_vis}
\vspace{-8pt}
\end{figure}

Fig.~\ref{fig_cls_vis_in_image} shows examples of the results from our frustum classification network and the baseline grid classification network (see supplementary materials). The point clouds are projected into the images using the ground truth pose $G$. The colors of the points represent the correctness of the frustum or grid predictions as described in the caption. The accuracy of the frustum and grid classifications are around 98\% and 51\% in Oxford, and 94\% and 39\% in KITTI. The low classification accuracy in KITTI leads to larger RTE and RRE during cross-modality registration.
3D visualization of the frustum classification and the inverse camera projection problem is illustrated in Fig.~\ref{fig_3d_vis}. It illustrates the intuition that a camera pose can be found by aligning the camera frustum to the classification result.

\section{Conclusions}
The paper proposes an approach for cross-modality registration between images and point clouds. The challenging registration problem is converted to a classification problem solved by deep networks and an inverse camera projection problem solved by least squares optimization. The feasibility of our proposed classification-optimization framework is verified with Oxford and KITTI dataset. 

\vspace{-8pt}
\paragraph{Acknowledgment.} \label{sec_ack}
This work is supported in part by the Singapore MOE Tier 1 grant R-252-000-A65-114.

\clearpage

{\small
\bibliographystyle{ieee_fullname}
\bibliography{main}

\begin{thebibliography}{10}\itemsep=-1pt

\bibitem{ceres-solver}
Sameer Agarwal, Keir Mierle, and Others.
\newblock Ceres solver.
\newblock \url{http://ceres-solver.org}.

\bibitem{besl1992method}
Paul~J Besl and Neil~D McKay.
\newblock Method for registration of 3-d shapes.
\newblock In {\em Sensor fusion IV: control paradigms and data structures},
  volume 1611, pages 586--606. International Society for Optics and Photonics,
  1992.

\bibitem{biber2003normal}
Peter Biber and Wolfgang Stra{\ss}er.
\newblock The normal distributions transform: A new approach to laser scan
  matching.
\newblock In {\em Proceedings 2003 IEEE/RSJ International Conference on
  Intelligent Robots and Systems (IROS 2003)(Cat. No. 03CH37453)}, volume~3,
  pages 2743--2748. IEEE, 2003.

\bibitem{cattaneo2020global}
Daniele Cattaneo, Matteo Vaghi, Simone Fontana, Augusto~Luis Ballardini, and
  Domenico~Giorgio Sorrenti.
\newblock Global visual localization in lidar-maps through shared 2d-3d
  embedding space.
\newblock In {\em 2020 IEEE International Conference on Robotics and Automation
  (ICRA)}, pages 4365--4371. IEEE, 2020.

\bibitem{chen1992object}
Yang Chen and G{\'e}rard Medioni.
\newblock Object modelling by registration of multiple range images.
\newblock {\em Image and vision computing}, 10(3):145--155, 1992.

\bibitem{deng2018ppf}
Haowen Deng, Tolga Birdal, and Slobodan Ilic.
\newblock Ppf-foldnet: Unsupervised learning of rotation invariant 3d local
  descriptors.
\newblock In {\em Proceedings of the European Conference on Computer Vision
  (ECCV)}, pages 602--618, 2018.

\bibitem{deng2018ppfnet}
Haowen Deng, Tolga Birdal, and Slobodan Ilic.
\newblock Ppfnet: Global context aware local features for robust 3d point
  matching.
\newblock In {\em Proceedings of the IEEE Conference on Computer Vision and
  Pattern Recognition}, pages 195--205, 2018.

\bibitem{dorai1997cosmos}
Chitra Dorai and Anil~K. Jain.
\newblock Cosmos-a representation scheme for 3d free-form objects.
\newblock {\em IEEE Transactions on Pattern Analysis and Machine Intelligence},
  19(10):1115--1130, 1997.

\bibitem{engel2017direct}
Jakob Engel, Vladlen Koltun, and Daniel Cremers.
\newblock Direct sparse odometry.
\newblock {\em IEEE transactions on pattern analysis and machine intelligence},
  40(3):611--625, 2017.

\bibitem{engel2014lsd}
Jakob Engel, Thomas Sch{\"o}ps, and Daniel Cremers.
\newblock Lsd-slam: Large-scale direct monocular slam.
\newblock In {\em European conference on computer vision}, pages 834--849.
  Springer, 2014.

\bibitem{feng20192d3d}
Mengdan Feng, Sixing Hu, Marcelo~H Ang, and Gim~Hee Lee.
\newblock 2d3d-matchnet: Learning to match keypoints across 2d image and 3d
  point cloud.
\newblock In {\em 2019 International Conference on Robotics and Automation
  (ICRA)}, pages 4790--4796. IEEE, 2019.

\bibitem{fischler1981random}
Martin~A Fischler and Robert~C Bolles.
\newblock Random sample consensus: a paradigm for model fitting with
  applications to image analysis and automated cartography.
\newblock {\em Communications of the ACM}, 24(6):381--395, 1981.

\bibitem{geiger2013vision}
Andreas Geiger, Philip Lenz, Christoph Stiller, and Raquel Urtasun.
\newblock Vision meets robotics: The kitti dataset.
\newblock {\em The International Journal of Robotics Research},
  32(11):1231--1237, 2013.

\bibitem{godard2019digging}
Cl{\'e}ment Godard, Oisin Mac~Aodha, Michael Firman, and Gabriel~J Brostow.
\newblock Digging into self-supervised monocular depth estimation.
\newblock In {\em Proceedings of the IEEE international conference on computer
  vision}, pages 3828--3838, 2019.

\bibitem{gojcic2019perfect}
Zan Gojcic, Caifa Zhou, Jan~D Wegner, and Andreas Wieser.
\newblock The perfect match: 3d point cloud matching with smoothed densities.
\newblock In {\em Proceedings of the IEEE Conference on Computer Vision and
  Pattern Recognition}, pages 5545--5554, 2019.

\bibitem{hartley2003multiple}
Richard Hartley and Andrew Zisserman.
\newblock {\em Multiple view geometry in computer vision}.
\newblock Cambridge university press, 2003.

\bibitem{he2016deep}
Kaiming He, Xiangyu Zhang, Shaoqing Ren, and Jian Sun.
\newblock Deep residual learning for image recognition.
\newblock In {\em Proceedings of the IEEE conference on computer vision and
  pattern recognition}, pages 770--778, 2016.

\bibitem{hess2016real}
Wolfgang Hess, Damon Kohler, Holger Rapp, and Daniel Andor.
\newblock Real-time loop closure in 2d lidar slam.
\newblock In {\em 2016 IEEE International Conference on Robotics and Automation
  (ICRA)}, pages 1271--1278. IEEE, 2016.

\bibitem{kneip2014upnp}
Laurent Kneip, Hongdong Li, and Yongduek Seo.
\newblock Upnp: An optimal o (n) solution to the absolute pose problem with
  universal applicability.
\newblock In {\em European Conference on Computer Vision}, pages 127--142.
  Springer, 2014.

\bibitem{lepetit2009epnp}
Vincent Lepetit, Francesc Moreno-Noguer, and Pascal Fua.
\newblock Epnp: An accurate o (n) solution to the pnp problem.
\newblock {\em International journal of computer vision}, 81(2):155, 2009.

\bibitem{li2018so}
Jiaxin Li, Ben~M Chen, and Gim Hee~Lee.
\newblock So-net: Self-organizing network for point cloud analysis.
\newblock In {\em Proceedings of the IEEE conference on computer vision and
  pattern recognition}, pages 9397--9406, 2018.

\bibitem{li2019usip}
Jiaxin Li and Gim~Hee Lee.
\newblock Usip: Unsupervised stable interest point detection from 3d point
  clouds.
\newblock In {\em Proceedings of the IEEE International Conference on Computer
  Vision}, pages 361--370, 2019.

\bibitem{lowe1999object}
David~G Lowe.
\newblock Object recognition from local scale-invariant features.
\newblock In {\em Proceedings of the seventh IEEE international conference on
  computer vision}, volume~2, pages 1150--1157. Ieee, 1999.

\bibitem{lu2019deepicp}
Weixin Lu, Guowei Wan, Yao Zhou, Xiangyu Fu, Pengfei Yuan, and Shiyu Song.
\newblock Deepicp: An end-to-end deep neural network for 3d point cloud
  registration.
\newblock {\em arXiv preprint arXiv:1905.04153}, 2019.

\bibitem{maddern20171}
Will Maddern, Geoffrey Pascoe, Chris Linegar, and Paul Newman.
\newblock 1 year, 1000 km: The oxford robotcar dataset.
\newblock {\em The International Journal of Robotics Research}, 36(1):3--15,
  2017.

\bibitem{mur2015orb}
Raul Mur-Artal, Jose Maria~Martinez Montiel, and Juan~D Tardos.
\newblock Orb-slam: a versatile and accurate monocular slam system.
\newblock {\em IEEE transactions on robotics}, 31(5):1147--1163, 2015.

\bibitem{pham2020lcd}
Quang-Hieu Pham, Mikaela~Angelina Uy, Binh-Son Hua, Duc~Thanh Nguyen, Gemma
  Roig, and Sai-Kit Yeung.
\newblock Lcd: learned cross-domain descriptors for 2d-3d matching.
\newblock In {\em Proceedings of the AAAI Conference on Artificial
  Intelligence}, volume~34, pages 11856--11864, 2020.

\bibitem{qi2017pointnet}
Charles~R Qi, Hao Su, Kaichun Mo, and Leonidas~J Guibas.
\newblock Pointnet: Deep learning on point sets for 3d classification and
  segmentation.
\newblock In {\em Proceedings of the IEEE conference on computer vision and
  pattern recognition}, pages 652--660, 2017.

\bibitem{qi2017pointnet++}
Charles~Ruizhongtai Qi, Li Yi, Hao Su, and Leonidas~J Guibas.
\newblock Pointnet++: Deep hierarchical feature learning on point sets in a
  metric space.
\newblock In {\em Advances in neural information processing systems}, pages
  5099--5108, 2017.

\bibitem{rublee2011orb}
Ethan Rublee, Vincent Rabaud, Kurt Konolige, and Gary Bradski.
\newblock Orb: An efficient alternative to sift or surf.
\newblock In {\em 2011 International conference on computer vision}, pages
  2564--2571. Ieee, 2011.

\bibitem{rusu2009fast}
Radu~Bogdan Rusu, Nico Blodow, and Michael Beetz.
\newblock Fast point feature histograms (fpfh) for 3d registration.
\newblock In {\em 2009 IEEE international conference on robotics and
  automation}, pages 3212--3217. IEEE, 2009.

\bibitem{rusu20113d}
Radu~Bogdan Rusu and Steve Cousins.
\newblock 3d is here: Point cloud library (pcl).
\newblock In {\em 2011 IEEE international conference on robotics and
  automation}, pages 1--4. IEEE, 2011.

\bibitem{sattler2012improving}
Torsten Sattler, Bastian Leibe, and Leif Kobbelt.
\newblock Improving image-based localization by active correspondence search.
\newblock In {\em European conference on computer vision}, pages 752--765.
  Springer, 2012.

\bibitem{shavit2019introduction}
Yoli Shavit and Ron Ferens.
\newblock Introduction to camera pose estimation with deep learning.
\newblock {\em arXiv preprint arXiv:1907.05272}, 2019.

\bibitem{tombari2010unique}
Federico Tombari, Samuele Salti, and Luigi Di~Stefano.
\newblock Unique signatures of histograms for local surface description.
\newblock In {\em European conference on computer vision}, pages 356--369.
  Springer, 2010.

\bibitem{tombari2013performance}
Federico Tombari, Samuele Salti, and Luigi Di~Stefano.
\newblock Performance evaluation of 3d keypoint detectors.
\newblock {\em International Journal of Computer Vision}, 102(1-3):198--220,
  2013.

\bibitem{triggs1999bundle}
Bill Triggs, Philip~F McLauchlan, Richard~I Hartley, and Andrew~W Fitzgibbon.
\newblock Bundle adjustment—a modern synthesis.
\newblock In {\em International workshop on vision algorithms}, pages 298--372.
  Springer, 1999.

\bibitem{ullman1979interpretation}
Shimon Ullman.
\newblock The interpretation of structure from motion.
\newblock {\em Proceedings of the Royal Society of London. Series B. Biological
  Sciences}, 203(1153):405--426, 1979.

\bibitem{wang2019deep}
Yue Wang and Justin~M Solomon.
\newblock Deep closest point: Learning representations for point cloud
  registration.
\newblock In {\em Proceedings of the IEEE International Conference on Computer
  Vision}, pages 3523--3532, 2019.

\bibitem{yang2013go}
Jiaolong Yang, Hongdong Li, and Yunde Jia.
\newblock Go-icp: Solving 3d registration efficiently and globally optimally.
\newblock In {\em Proceedings of the IEEE International Conference on Computer
  Vision}, pages 1457--1464, 2013.

\bibitem{yew20183dfeat}
Zi~Jian Yew and Gim~Hee Lee.
\newblock 3dfeat-net: Weakly supervised local 3d features for point cloud
  registration.
\newblock In {\em European Conference on Computer Vision}, pages 630--646.
  Springer, 2018.

\bibitem{yew2020-RPMNet}
Zi~Jian Yew and Gim~Hee Lee.
\newblock Rpm-net: Robust point matching using learned features.
\newblock In {\em Conference on Computer Vision and Pattern Recognition
  (CVPR)}, 2020.

\bibitem{yu2020monocular}
Huai Yu, Weikun Zhen, Wen Yang, Ji Zhang, and Sebastian Scherer.
\newblock Monocular camera localization in prior lidar maps with 2d-3d line
  correspondences.
\newblock {\em arXiv preprint arXiv:2004.00740}, 2020.

\bibitem{zeng20173dmatch}
Andy Zeng, Shuran Song, Matthias Nie{\ss}ner, Matthew Fisher, Jianxiong Xiao,
  and Thomas Funkhouser.
\newblock 3dmatch: Learning local geometric descriptors from rgb-d
  reconstructions.
\newblock In {\em Proceedings of the IEEE Conference on Computer Vision and
  Pattern Recognition}, pages 1802--1811, 2017.

\bibitem{zhang2014loam}
Ji Zhang and Sanjiv Singh.
\newblock Loam: Lidar odometry and mapping in real-time.
\newblock In {\em Robotics: Science and Systems}, volume~2, 2014.

\bibitem{zhong2009intrinsic}
Yu Zhong.
\newblock Intrinsic shape signatures: A shape descriptor for 3d object
  recognition.
\newblock In {\em 2009 IEEE 12th International Conference on Computer Vision
  Workshops, ICCV Workshops}, pages 689--696. IEEE, 2009.

\end{thebibliography}
}

\clearpage

\appendix

\section{``Grid Classification + PnP" Method}
\subsection{Grid Classification}
We divide the $H\times W$ image into a tessellation of $32\times 32$ regions and then assign a label to each region. For example, an $128\times 512$ image effectively becomes $4\times 16=64$ patches, and the respective regions are assigned to take a label $l^f \in \{0, 1, \cdots, 63\}$. In the per-point classification, a point taking a label $l^f$ projects to the image region with the same label. Consequently, the grid classification is actually downsampling the image by a factor of 32, and reveals the pixel of the downsampled image to point correspondence. Formally, the grid classification assigns a label to each point, $L^f = \{l^f_1, l^f_2, \cdots, l^f_N\}$, where $l^f_n \in \{0, 1, \cdots, \frac{H\times W}{32\times 32}-1\}$.

\paragraph{Label Generation.}
Grid classification is performed only on points that are predicted as inside camera frustum, i.e. $\hat{l}_i=1$. The goal of grid classification is to get the assignment of the point to one of the $32\times 32$ patches. We define the labels from the grid classification as:
\begin{equation} \label{equ_grid_labels}
    l^f_i = \bigg\lfloor \frac{p'_{x_i}}{32} \bigg\rfloor + \bigg\lfloor \frac{p'_{y_i}}{32} \bigg\rfloor \cdot \frac{W}{32},
\end{equation}
where $\lfloor . \rfloor$ is the floor operator. Note that the image width and height $(W, H)$ are required to be a multiple of $32$.

\paragraph{Training the Grid Classifier.}
As mentioned in Section~4.2, the training of the grid classifier is very similar to the frustum classifier with the exception that the labels are different. Nonetheless, the frustum and grid classifier can be trained together as shown in Fig.~1.

\subsection{PnP} \label{sec_ransac_pnp}
Given the grid classifier results, the pose estimation problem can be formulated as a Perspective-n-Point (PnP) problem. The grid classification effectively builds correspondences between each point to the subsampled image, e.g. subsampled by 32. The PnP problem is to solve the rotation $R$ and translation $\mathbf{t}$ of the camera from a given set of 3D points $\{\mathbf{P}_1, \cdots, \mathbf{P}_{M} \mid \mathbf{P}_m \in \mathbb{R}^3\}$, the corresponding image pixels $\{\mathbf{p}_1, \cdots, \mathbf{p}_{M} \mid \mathbf{p}_m \in \mathbb{R}^2\}$, and the camera intrinsic matrix $K \in \mathbb{R}^{3\times 3}$.
The 3D points are those classified as within the image by the frustum classification, i.e. $P = \{\mathbf{P}_1, \cdots, \mathbf{P}_{M}\}$, where $\mathbf{P}_m$ is the point $\mathbf{P}_n \in P: \hat{l}^c_n = 1$. The corresponding pixels are acquired given by:
\begin{equation} \label{equ_pnp_pixels}
        p_{y_i} = \bigg\lfloor \frac{l^f_i}{W'}\bigg \rfloor, p_{x_i} = l^f_i - W' p_{y_i}, \text{\,\,where\,\,} W'=\frac{W}{32}, 
\end{equation}
and $L^f = \{l^f_1, \cdots, l^f_{M}\},~l^f_i \in [0, (HW/(32\times32))-1]$ is the prediction from the grid classification. 
We can effectively solve for the unknown pose $\hat{G} \in \text{SE}(3)$ in the PnP problem after  
resizing the image into $1/32$ of the original size.
Accordingly, the camera intrinsics $K'$ after the resize is obtained by dividing $f_x$, $f_y$, $c_x$, $c_y$ with $32$.
There are many off-the-shelf PnP solver like EPnP \cite{lepetit2009epnp}, UPnP \cite{kneip2014upnp}, etc. We apply RANSAC on EPnP provided by OpenCV to robustly solve for $\hat{G}$.

\paragraph{Implementation details.} The RANSAC PnP \cite{fischler1981random} from OpenCV does not require initialization. We set the threshold for inlier reprojection error to $0.6$ and maximum iteration number to $500$ in RANSAC. Note that we can optionally use the results from RANSAC PnP to initialize the inverse camera projection optimization. 

\subsection{Experiments} 

\paragraph{Inverse~Camera~Projection vs RANSAC~PnP \cite{fischler1981random}.} As shown in Table~1, the inverse camera projection solver with 3-DoF performs the best. This verifies the effectiveness of our solver design in Section~5. Nonetheless, the advantage of RANSAC PnP over the inverse camera projection solver is that it does not require initialization, and its performance with 6-DoF is also sufficiently good.

\section{Classification Network Details}

\subsection{Point Cloud Encoder}
\begin{figure}[h!] \centering
\subfigure[]{\includegraphics[height=0.16\textwidth]{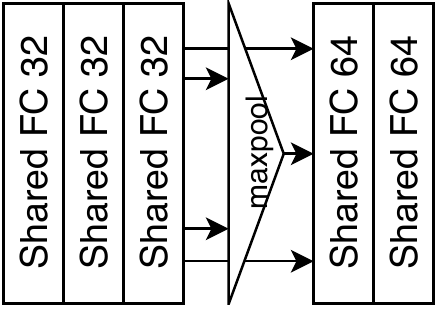}\label{fig_pn1}} \hspace{10mm} 
\subfigure[]{\includegraphics[height=0.16\textwidth]{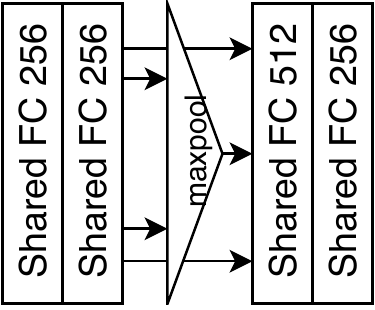}\label{fig_pn2}} \hspace{10mm}
\subfigure[]{\includegraphics[height=0.16\textwidth]{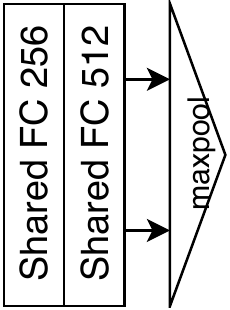}\label{fig_pn3}} \hspace{10mm}
\caption{Network details in Point Cloud Encoder. (a) (b) (c) are the PointNet-like network structures used in the encoder.}
\end{figure}

The input point cloud is randomly downsampled to a size of 20,480. The Lidar intensity values are appended to the x-y-z coordinates for each point. Consequently, the size of the input data is $4\times 20,480$. 
During the first sampling-grouping-PointNet operation, the FPS operation extracts $M_1=128$ nodes denoted as $\mathfrak{P}^{(1)}$. The grouping procedure is exactly the same as the point-to-node method described in SO-Net \cite{li2018so}. 
As shown in Fig.~\ref{fig_pn1}, our PointNet-like module, which produces the feature $P^{(1)}$, is a slight modification of the original PointNet \cite{qi2017pointnet}. At the second sampling-grouping-PointNet operation, the FPS extracts $M_2=64$ nodes denoted as $\mathfrak{P}^{(2)}$. The grouping step is a $k$NN-based operation as described in PointNet++ \cite{qi2017pointnet++}. Each node in $\mathfrak{P}^{(2)}$ are connected to its 16 nearest neighbors in $\mathfrak{P}^{(1)}$. The feature $P^{(2)}$ for each node in $\mathfrak{P}^{(2)}$ is obtained by the PointNet-like module shown in Fig.~\ref{fig_pn2}. Finally, a global point cloud feature vector is obtained by feeding $\mathfrak{P}^{(2)}$ and $P^{(2)}$ into a PointNet module shown in Fig.~\ref{fig_pn3}

\subsection{Image-Point Cloud Attention Fusion}
The first attention fusion module takes the image features $I^{(1)}\in \mathbb{R}^{256\times H_1\times W_1}, H_1 = H/16, W_1=W/16$, global image feature $I^{(3)}\in \mathbb{R}^{512}$, and point cloud feature $P^{(1)}\in \mathbb{R}^{C_1\times M_1}$ as input. A shared MLP takes $I^{(3)}, P^{(1)}$ as input and produces the weighting $S_{att}^{(1)}\in \mathbb{R}^{(H_1\cdot W_1) \times M_1}$. The shared MLP consists of two fully connected layers. The weighted image feature $\tilde{I}^{(1)} \in \mathbb{R}^{256\times M_1}$ is from the multiplication of $I^{(1)}$ with $S_{att}^{(1)}$. $\tilde{I}^{(1)}$ is then used in the Point Cloud Decoder.
Similarly, $\tilde{I}^{(2)} \in \mathbb{R}^{512\times M_1}$ is acquired using shared MLP of the same structure, which takes $I^{(3)}, P^{(2)}$ as input; and outputs the weighting $S_{att}^{(2)}\in \mathbb{R}^{(H_2\cdot W_2) \times M_2}$.

\subsection{Point Cloud Decoder}
\begin{figure}[h!] \centering
\subfigure[]{\includegraphics[height=0.16\textwidth]{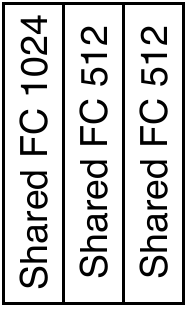}\label{fig_mlp1}} \hspace{10mm} 
\subfigure[]{\includegraphics[height=0.16\textwidth]{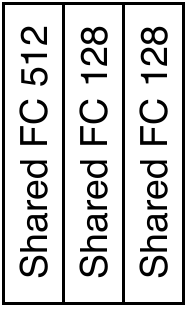}\label{fig_mlp2}} \hspace{10mm}
\subfigure[]{\includegraphics[height=0.16\textwidth]{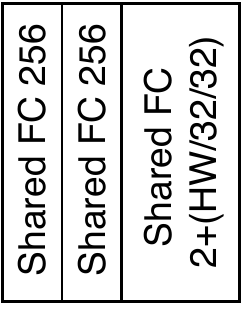}\label{fig_mlp3}} \hspace{10mm}
\caption{Network details in Point Cloud Decoder. (a) (b) (c) are the shared MLPs used in the encoder.}
\end{figure}

There are two concatenate-sharedMLP-interpolation processes in the decoder to get $\tilde{P}^{(2)}_{(itp)}\in \mathbb{R}^{C_2\times M_1}$ and $\tilde{P}^{(1)}_{(itp)}\in \mathbb{R}^{C_1\times N}$. In both interpolation operations, the $k$ nearest neighbor search is configured as $k=16$. The shared MLP that takes $[I^{(3)}, \tilde{I}^{(2)}, P^{(3)}, P^{(2)}]$ to produce $\tilde{P}^{(2)} \in \mathbb{R}^{C_2\times M_2}$ is shown in Fig.~\ref{fig_mlp1}. 
Similarly, the shared MLP that takes $[\tilde{P}^{(2)}_{(itp)}, \tilde{I}^{(1)}]$ to produce $\tilde{P}^{(1)} \in \mathbb{R}^{C_1\times M_1}$ is shown in Fig.~\ref{fig_mlp2}. Finally, the shared MLP shown in Fig.~\ref{fig_mlp3} takes $[P^{(1)}, \tilde{P}^{(1)}_{(itp)}\in \mathbb{R}^{C_1\times N}]$ to produce the frustum and grid predictions scores.

\section{Experiment Details}
\begin{figure*}[t!] \centering
{\includegraphics[width=0.90\textwidth]{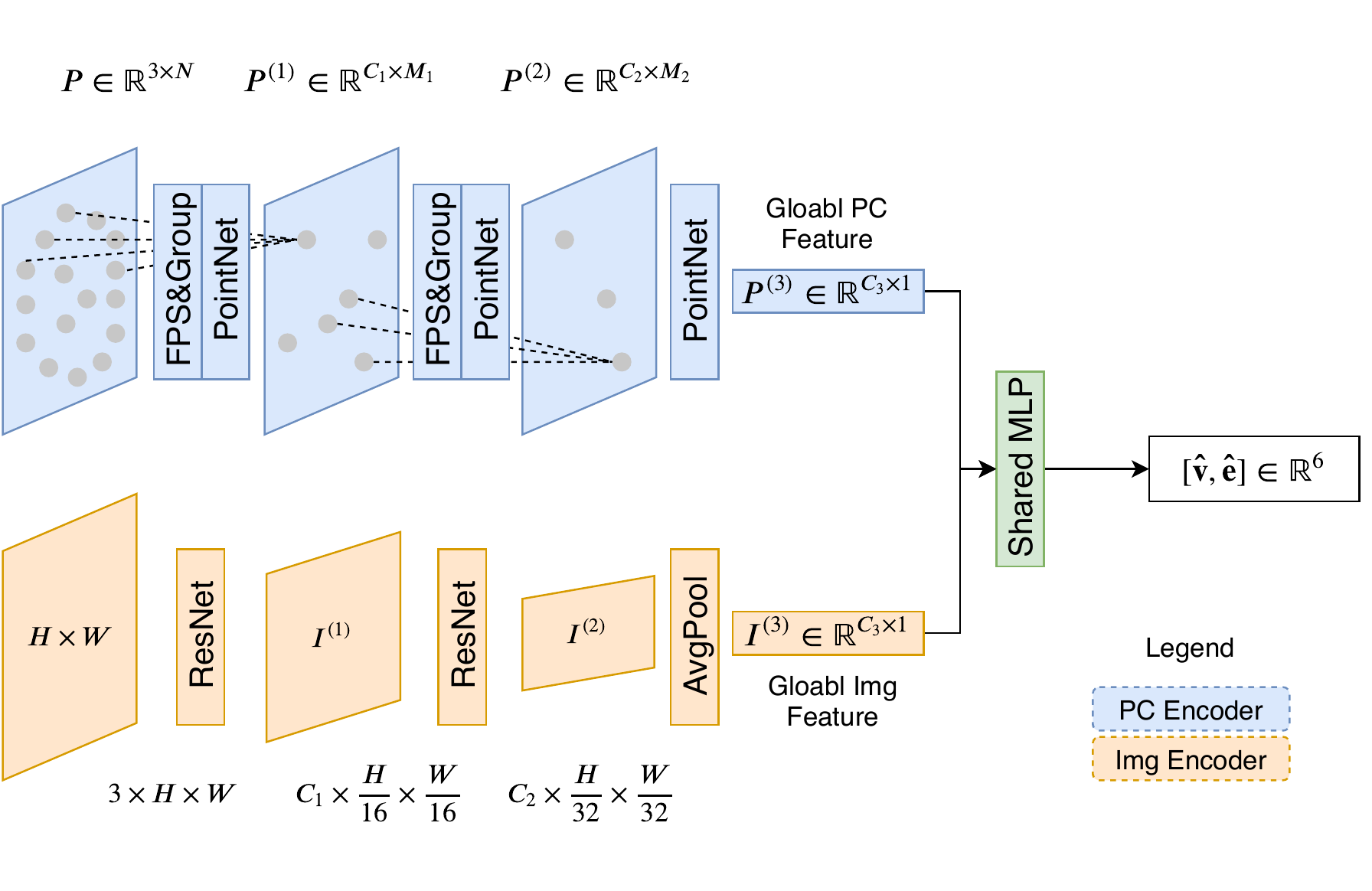}}
\caption{Our network architecture for the baseline method.} \label{fig_baseline}
\end{figure*}
\subsection{Dataset Configurations}
In the Oxford dataset, the point clouds are built from the accumulation of the 2D scans from a 2D Lidar. Each point cloud is set at the size of radius $50$m, i.e. diameter $100$m. Point clouds are built every $2$m to get 130,078 point clouds for training and 19,156 for testing. There are a lot more training/testing images because they are randomly sampled within $\pm10$m. Note that we do not use night driving traversals for training and testing because the image quality at night is too low for cross-modality registration.
The images are captured by the center camera of a Bumblebee tri-camera rig. The bottom 160 rows of the image is cropped out because those rows are occupied by the egocar. The $800\times 1280$ image is resized to $400\times 640$ and then random/center cropped into $384\times 640$ during training/testing. 

In KITTI Odometry dataset, point clouds are directly acquired from a 3D Lidar. Every point cloud in the dataset is used for either training or testing. We follow the common practice of utilizing the 0-8 sequences for training, and 9-10 for testing. In total there are 20,409 point clouds for training, and 2,792 for testing. 
The top 100 rows of the images are cropped out because they are mostly seeing the sky. The original $320\times 1224$ images are resized into $160\times 612$, and then random/center cropped into $160\times 512$ during training/testing.

\subsection{``Direct Regression" Method}
The direct regression method is a deep network-based approach that directly regresses the pose between a pair of image and point cloud. The network architecture is shown in Fig.~\ref{fig_baseline}. The Point Cloud Encoder and Image Encoder are exactly the same as the classification network in our DeepI2P. The global point cloud feature $P^{(3)} \in \mathbb{R}^{512}$ and global image feature $I^{(3)}\in \mathbb{R}^{512}$ are fed into a MLP to produce the relative pose. The relative translation is represented by a vector $\mathbf{\hat{v}} \in \mathbb{R}^3$, while the relative rotation is represented by angle-axis $\mathbf{\hat{e}}\in \mathbb{R}^3$. Given the ground truth $\mathbf{v}\in \mathbb{R}^{3}$ and rotation $R\in \mathbb{R}^{3\times 3}$, the loss function is given by:
\begin{equation} \label{equ_baseline_loss}
    \mathcal{L} = \mathcal{L}_{tran} + \mathcal{L}_{rot} = \|\mathbf{v}-\mathbf{\hat{v}}\|_2 + \|f(\mathbf{\hat{e}}) - R\|_F,
\end{equation}
where $f(\cdot)$ is the funtion that converts the angle-axis representation $\mathbf{\hat{e}}\in \mathbb{R}^3$ to a rotation matrix $\hat{R}\in \mathbb{R}^{3\times 3}$, and $\|\cdot\|_F$ is the matrix Frobenius norm.
The training configurations are the same as our DeepI2P, i.e. the image and point cloud are within $10$m and additional random 2D rotation is applied to the point cloud.

\end{document}